\documentclass[10pt,twocolumn,letterpaper]{article}

\usepackage[cvprfinal]{cvpr}      
\usepackage[accsupp]{axessibility}

\usepackage{graphicx}
\usepackage{amsmath}
\usepackage{amssymb}
\usepackage{booktabs}

\usepackage[pagebackref=true,breaklinks=true,colorlinks=true]{hyperref}

\usepackage[capitalize]{cleveref}
\crefname{section}{Sec.}{Secs.}
\Crefname{section}{Section}{Sections}
\Crefname{table}{Table}{Tables}
\crefname{table}{Tab.}{Tabs.}

\def\cvprPaperID{*****} 
\def\confName{CVPR}
\def\confYear{2024}

\begin{document}

\title{VMRNN: Integrating Vision Mamba and LSTM for Efficient and Accurate Spatiotemporal Forecasting} 

\author{Yujin Tang$^{1}$\quad Peijie Dong$^{2}$\quad Zhenheng Tang$^{2,3}$\quad Xiaowen Chu$^{2}$\quad Junwei Liang$^{1,4}$ \thanks{Corresponding author} \\
$^{1}$AI Thrust, The Hong Kong University of Science and Technology (Guangzhou) \quad \\$^{2}$DSA Thrust, The Hong Kong University of Science and Technology (Guangzhou)\\$^{3}$Department of Computer Science, Hong Kong Baptist University\\$^{4}$Department of Computer Science and Engineering, The Hong Kong University of Science and Technology \\
\small{\texttt{tangyujin0275@gmail.com,junweiliang@hkust-gz.edu.cn}}}

\maketitle

\begin{abstract}

Combining Convolutional Neural Networks (CNNs) or Vision Transformers(ViTs) with Recurrent Neural Networks (RNNs) for spatiotemporal forecasting has yielded unparalleled results in predicting temporal and spatial dynamics. However, modeling extensive global information remains a formidable challenge; CNNs are limited by their narrow receptive fields, and ViTs struggle with the intensive computational demands of their attention mechanisms. The emergence of recent Mamba-based architectures has been met with enthusiasm for their exceptional long-sequence modeling capabilities, surpassing established vision models in efficiency and accuracy, which motivates us to develop an innovative architecture tailored for spatiotemporal forecasting. In this paper, we propose the VMRNN cell, a new recurrent unit that integrates the strengths of Vision Mamba blocks with LSTM. We construct a network centered on VMRNN cells to tackle spatiotemporal prediction tasks effectively. Our extensive evaluations show that our proposed approach secures competitive results on a variety of tasks while maintaining a smaller model size. Our code is available at \href{https://github.com/yyyujintang/VMRNN-PyTorch}{https://github.com/yyyujintang/VMRNN-PyTorch}.
\end{abstract}

\section{Introduction}
\label{sec:intro}

\begin{figure}[htp]
	\centering
	\resizebox{0.5\textwidth}{!}{
	\includegraphics{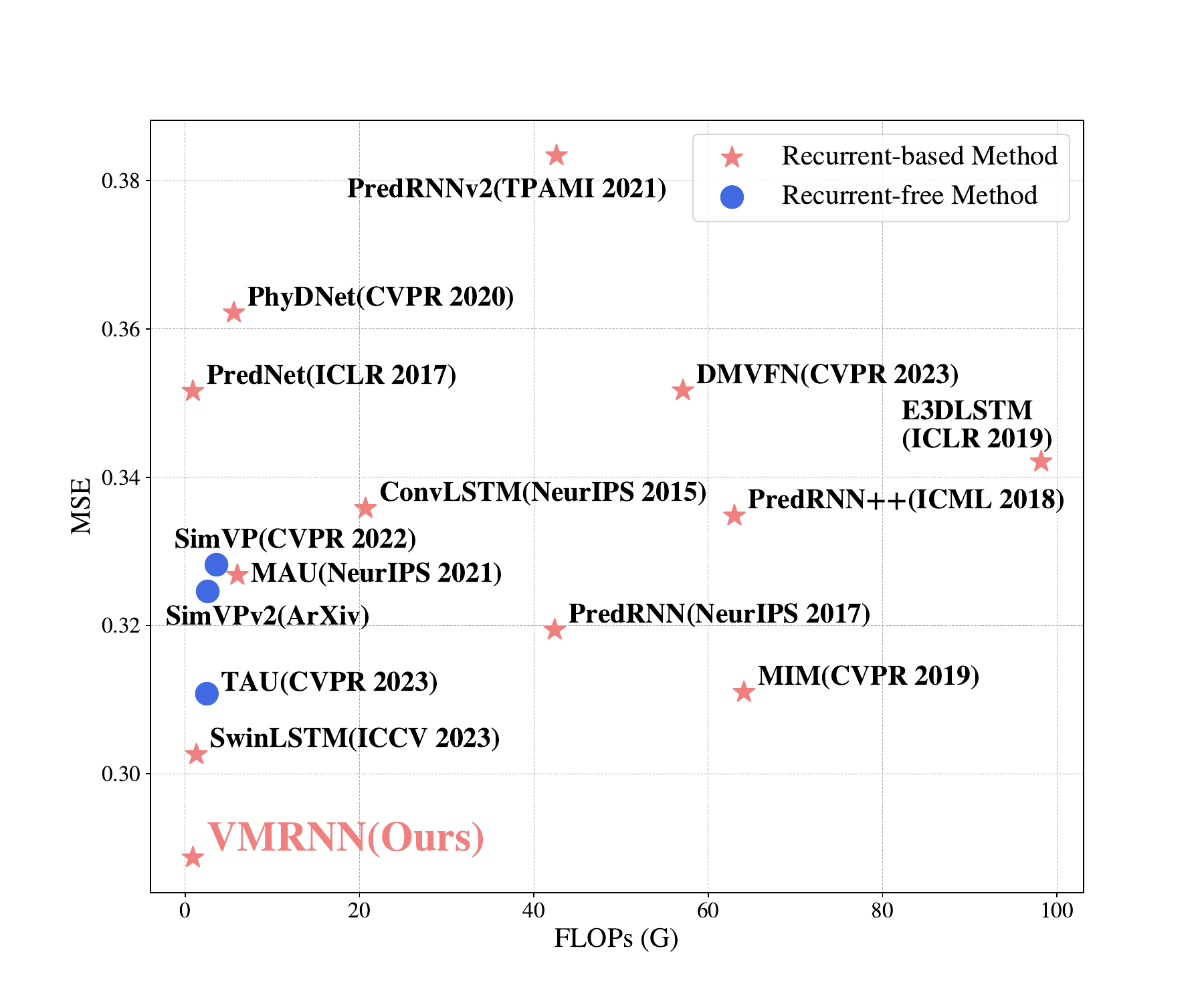}
	}
	\caption{Performance comparison on TaxiBJ over spatial-temporal predictive learning methods. VMRNN outperforms previous methods in terms of Mean-Squared-Error (MSE, lower the better) with a lower computational cost (GFLOPs).}
	\label{fig:flops}
\end{figure}

In recent years, spatiotemporal prediction has experienced a surge in interest due to its potential to enhance a wide range of practical applications. These applications span from precipitation forecasting~\cite{shi2015convolutional,shi2017deep,wang2017predrnn,tang2023postrainbench}, autonomous driving~\cite{bhattacharyya2018long,kwon2019predicting,li2023tfnet}, traffic flow prediction~\cite{zhang2017deep,xu2018predcnn}, and human motion forecasting~\cite{zhang2017learning,wang2018rgb} to representation learning~\cite{qian2021spatiotemporal,jenni2020video}. The ability to accurately predict spatial and temporal variations holds immense promise for improving decision-making processes and operational efficiencies across diverse sectors, underscoring the importance of continued research and development in spatiotemporal analysis. The complex physical interactions and the unpredictable characteristics of spatiotemporal data present significant obstacles for solely data-driven deep learning approaches to attain accurate predictions. The essence of spatiotemporal predictive learning lies in its capability to delve into the spatial correlations and temporal progressions inherent in the physical realm, highlighting its potential to uncover deep insights into the dynamics of our world.

To address these challenges, a plethora of methodologies have been developed, including recurrent-based methods which combine Convolutional Neural Networks (CNNs) or Vision Transformers(ViTs) with Recurrent Neural Networks (RNNs)~\cite{shi2015convolutional,wang2017predrnn,wang2018predrnn++,wang2019memory,lee2021video,wang2018eidetic,chang2021mau,guen2020disentangling,yu2020efficient,Tang_2023_ICCV} and recurrent-free methods like SimVP~\cite{gao2022simvp,tan2022simvpv2}, which fully based on CNN. 
%
As for recurrent-based methods, lots of innovative RNNs are proposed. Among these, ConvLSTM~\cite{shi2015convolutional} represents a pivotal advancement by augmenting the fully connected LSTM with convolutional operations to simultaneously capture spatial and temporal dependencies. Building upon ConvLSTM, a variety of innovative approaches have emerged. PredRNN~\cite{wang2017predrnn} and MIM~\cite{wang2019memory}, for instance, refine the LSTM unit's internal mechanics, while E3D-LSTM~\cite{wang2018eidetic} introduces 3D convolutions into LSTM structures. PhyDNet utilizes a CNN-based approach to untangle physical dynamics, and MAU~\cite{chang2021mau} introduces a motion-aware unit for enhanced motion capture. Recent recurrent-free models, like SimVP~\cite{gao2022simvp}, and the Temporal Attention Unit (TAU)~\cite{tan2023temporal}, which bifurcates temporal attention into static intra-frame and dynamic inter-frame components, offer fresh perspectives on spatiotemporal modeling. Furthermore, advancements such as the Dynamic Multi-scale Voxel Flow Network (DMVFN)~\cite{hu2023dynamic} and the two-stream MMVP~\cite{hu2023dynamic} framework underscore the recent innovation in this field, emphasizing the separation of motion and appearance for improved prediction. SwinLSTM~\cite{Tang_2023_ICCV} successfully integrates Swin Transformer~\cite{liu2021swin} with LSTM which stands out as a strong spatiotemporal prediction baseline.

While these approaches have demonstrated notable success in spatiotemporal forecasting, CNNs are intrinsically constrained by their local receptive fields ~\cite{luo2016understanding}, limiting their capacity to assimilate information from distant image regions. ViTs generally exhibit superior performance compared to CNNs, which could be attributed to global receptive fields and dynamic weights facilitated by the attention mechanism. However, the attention mechanism requires quadratic complexity in terms of image sizes, resulting in expensive computational overhead when addressing downstream dense prediction tasks.

\begin{figure*}[htp]
	\centering
	\resizebox{\textwidth}{!}{
	\includegraphics{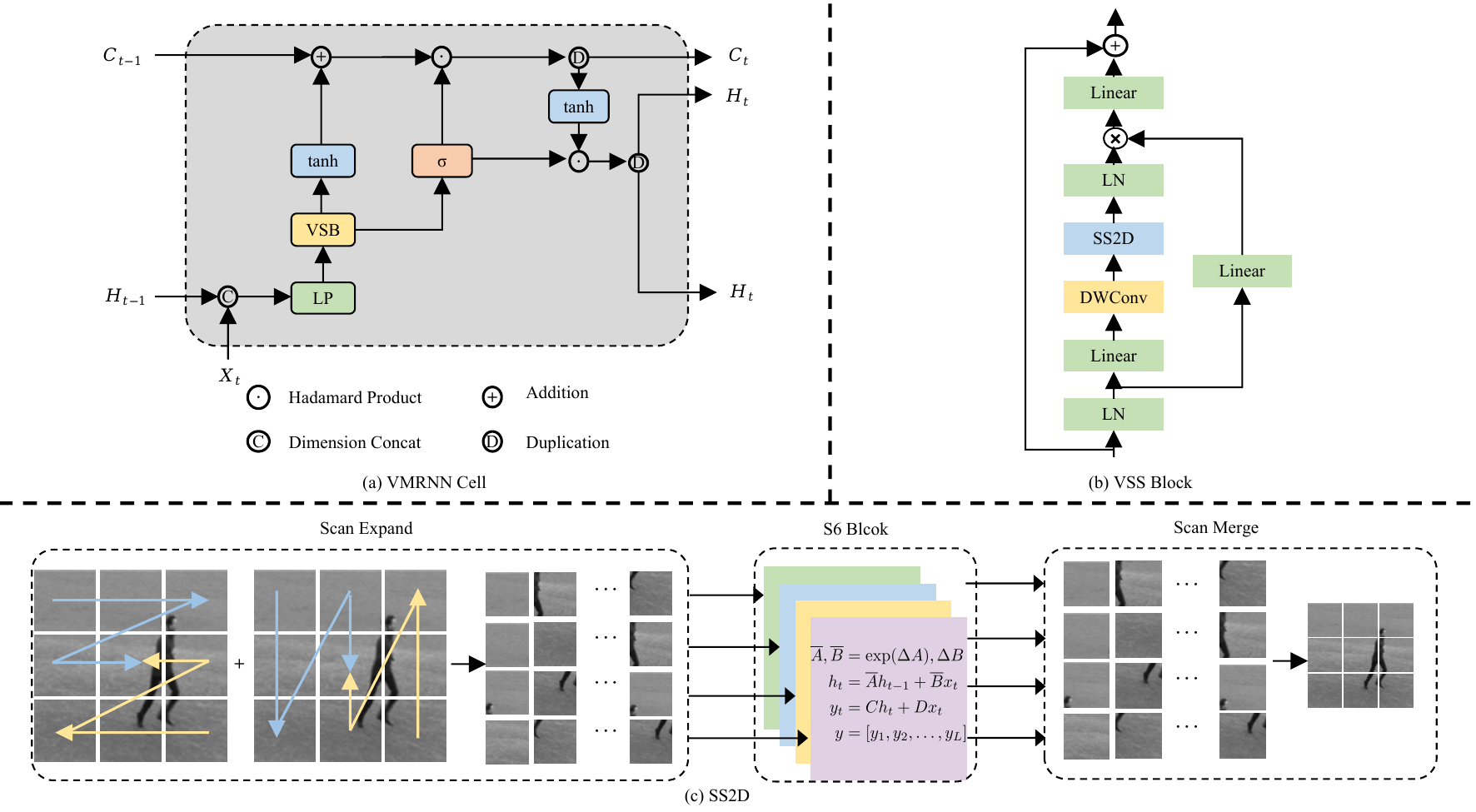}
	}
	\caption{(a): The detailed structure of the proposed recurrent cell: VMRNN. $\textbf{VSB}$ and $\textbf{LP}$ denote VSS Block and Linear Projection. (b): The architecture of VSS Block. (c): The SS2D process, includes three stages: Scan Expand, S6 Block, and Scan Merge.}
	\label{fig:VMRNNCell}
\end{figure*}

Recently, Structured State Space Models (S4) ~\cite{s4gu2021,gu2021combining} have emerged as notably efficient and effective in modeling extensive sequences. Mamba~\cite{mamba} positions itself as a potential breakthrough for addressing long-range dependencies in various tasks, which innovatively introduces the selectivity of the input sequence and uses the scan method. Unlike transformers, which often exhibit quadratic scaling for sequence length, Mamba maintains a linear or near-linear scaling, all the while adeptly handling long-range dependencies. This attribute has catapulted them to the forefront of continuous long-sequence data analysis, achieving state-of-the-art results in fields such as natural language processing and genomic analysis. A series of pioneering studies~\cite{umamba,liu2024vmamba,segmamba,vim,guo2024mambamorph,liu2024swin,yang2024vivim,gong2024nnmamba,guo2024mambair} have begun to investigate the utility of Mamba models within the vision sector, showcasing their versatility across a broad spectrum of vision-based applications. These investigations into Mamba’s application—ranging from basic image recognition to complex segmentation tasks—have yielded encouraging outcomes, underscoring the significant potential that SSMs hold in revolutionizing vision tasks.

Inspired by these studies, we introduce VMRNN, an innovative recurrent cell that merges Vision Mamba blocks (VSS Block) with an LSTM module to effectively distill spatiotemporal representations. Furthermore, we develop a model centered around VMRNN, specifically engineered to discern both spatial and temporal dynamics crucial for spatiotemporal forecasting. Unlike previous image-level vision tasks, spatiotemporal predictive learning predicts future frames from past frames at the video level. Our model processes each frame at the image level, segments them into patches, and flattens these patches before passing them to the patch embedding layer for preliminary processing. Our method inherits the attribute of the recurrent-based methods. The VMRNN layer utilizes these transformed patches with previous states to capture spatiotemporal representations for the next prediction.

The contributions of this research are threefold:
\begin{itemize}
	\item We introduce VMRNN, an innovative recurrent cell that fuses an LSTM architecture with Vision Mamba Blocks. To the best of our knowledge, we are the first to introduce Mamba into vision-based spatial-temporal forecasting to harness robust sequence modeling prowess.

	\item We propose two new architectures based on VMRNN, VMRNN-B, and VMRNN-D, excelling in extracting spatiotemporal representations and providing a new strong baseline for spatiotemporal forecasting.

	\item Extensive evaluation on Moving MNIST, TaxiBJ, and KTH demonstrates that our VMRNN not only shows a significant reduction in computational complexity and parameters but also matches or surpasses SOTA methods on all three datasets across metrics, validating its efficacy on three pivotal datasets.
\end{itemize}

\section{Related Work}

\subsection{Convolution-based Architecture}

Previous models that merge CNNs with RNNs adopt a range of tactics to more effectively grasp the nuances of spatiotemporal relationships, aiming to enhance predictive precision. ConvLSTM~\cite{shi2015convolutional} evolves from FC-LSTM~\cite{srivastava2015unsupervised} by integrating convolutional operations instead of fully connected ones, facilitating the learning of spatiotemporal interdependencies. PredRNN~\cite{wang2017predrnn} and its Spatiotemporal LSTM (ST-LSTM) unit mark a significant step, enabling the concurrent processing of spatiotemporal data by propagating hidden states both horizontally and vertically. Building on this, PredRNN++\cite{wang2018predrnn++} contributes the Gradient Highway unit to mitigate the vanishing gradient issue encountered by its predecessor. Meanwhile, E3D-LSTM\cite{wang2018eidetic} enhances the ST-LSTM's memory capacity by implementing 3D convolutions. The MIM model~\cite{wang2019memory} reimagines the ST-LSTM's forget gate with dual recurrent units to better address the learning of non-stationary information within predictions. CrevNet~\cite{yu2020efficient} employs a CNN-based reversible architecture to decode complex spatiotemporal patterns. Additionally, PhyDNet~\cite{guen2020disentangling} embeds physical principles into CNN frameworks to refine the quality of its predictions. Collectively, these models~\cite{shi2015convolutional,wang2017predrnn,wang2018predrnn++,wang2018eidetic,wang2019memory,yu2020efficient,guen2020disentangling} showcase a variety of approaches to enhance the capture of spatiotemporal dependencies and have garnered commendable outcomes. Nonetheless, conventional convolutional methodologies are constrained in their ability to seize spatiotemporal dependencies due to their intrinsic localized operation.

\subsection{Transformer-based Architecture}
The adoption of the Transformer model~\cite{vaswani2017attention}, initially celebrated in natural language processing, has prompted its exploration within the realm of computer vision. The Vision Transformer (ViT)\cite{dosovitskiy2020image} broke new ground by directly applying Transformer architecture to image classification, demonstrating impressive results. Further advancing this domain, the Swin Transformer\cite{liu2021swin} delivers remarkable achievements across a spectrum of tasks such as image classification, semantic segmentation, and object detection, thanks to its innovative shifted window strategy and hierarchical structure. Building on this, SwinLSTM~\cite{Tang_2023_ICCV} innovatively merges the Swin Transformer~\cite{liu2021swin} with LSTM, establishing a new robust benchmark for spatiotemporal forecasting. However, ViT and its derivatives exhibit a notable drawback: the attention mechanism's quadratic complexity in relation to image size, which imposes considerable computational demands.

\subsection{State Space Models}
State Space Models(SSMs) are recently proposed models that are introduced into deep learning as state space transforming~\cite{gu2021combining, s4gu2021, s5smith2022simplified}. 
Inspired by continuous state space models in control systems, combined with HiPPO~\cite{hippogu2020} initialization, LSSL~\cite{gu2021combining} showcases the potential to handle long-range dependency problems.
However, due to the prohibitive computation and memory requirements induced by the state representation, LSSL is infeasible to use in practice.
To solve this problem, S4 ~\cite{s4gu2021} proposes to normalize the parameter into the diagonal structure. Since then, many flavors of structured state space models sprang up with different structures like complex-diagonal structure ~\cite{dssgupta2022, s4dgu2022},  multiple-input multiple-output supporting ~\cite{s5smith2022simplified}, decomposition of diagonal plus low-rank operations ~\cite{liquids4hasani2022liquid}, selection mechanism ~\cite{mamba}. These models are then integrated into large representation models ~\cite{gssmehta2023long, megama2022mega, h3fu2022hungry}. Among these developments, Mamba~\cite{mamba} proposes the selective scan space state sequential model (S6) Block, which stands out as a promising innovation for tackling long-range dependencies across a spectrum of tasks. It introduces a novel approach by selectively processing the input sequence and employing a scanning method, marking a potential breakthrough in the field.

Several latest studies have preliminarily explored the effectiveness of Mamba in the vision domain. For instance, Vim ~\cite{vim} proposed a generic vision backbones with bidirectional Mamba blocks. In contrast, VMamba ~\cite{liu2024vmamba} builds up a Mamba-based vision backbone with hierarchical representations. Additionally, VMamba introduced a cross-scan module to solve the direction-sensitive problem due to the difference between 1D sequences and 2D images. In this paper, we try to integrate the Vision Mamba blocks proposed in VMamba with the simplified LSTM to form a VMRNN recurrent cell and use it as the core to build a model to capture temporal and spatial dependencies to perform spatiotemporal prediction tasks.

\begin{figure*}[htp]
	\centering
	\resizebox{\textwidth}{!}{
	\includegraphics{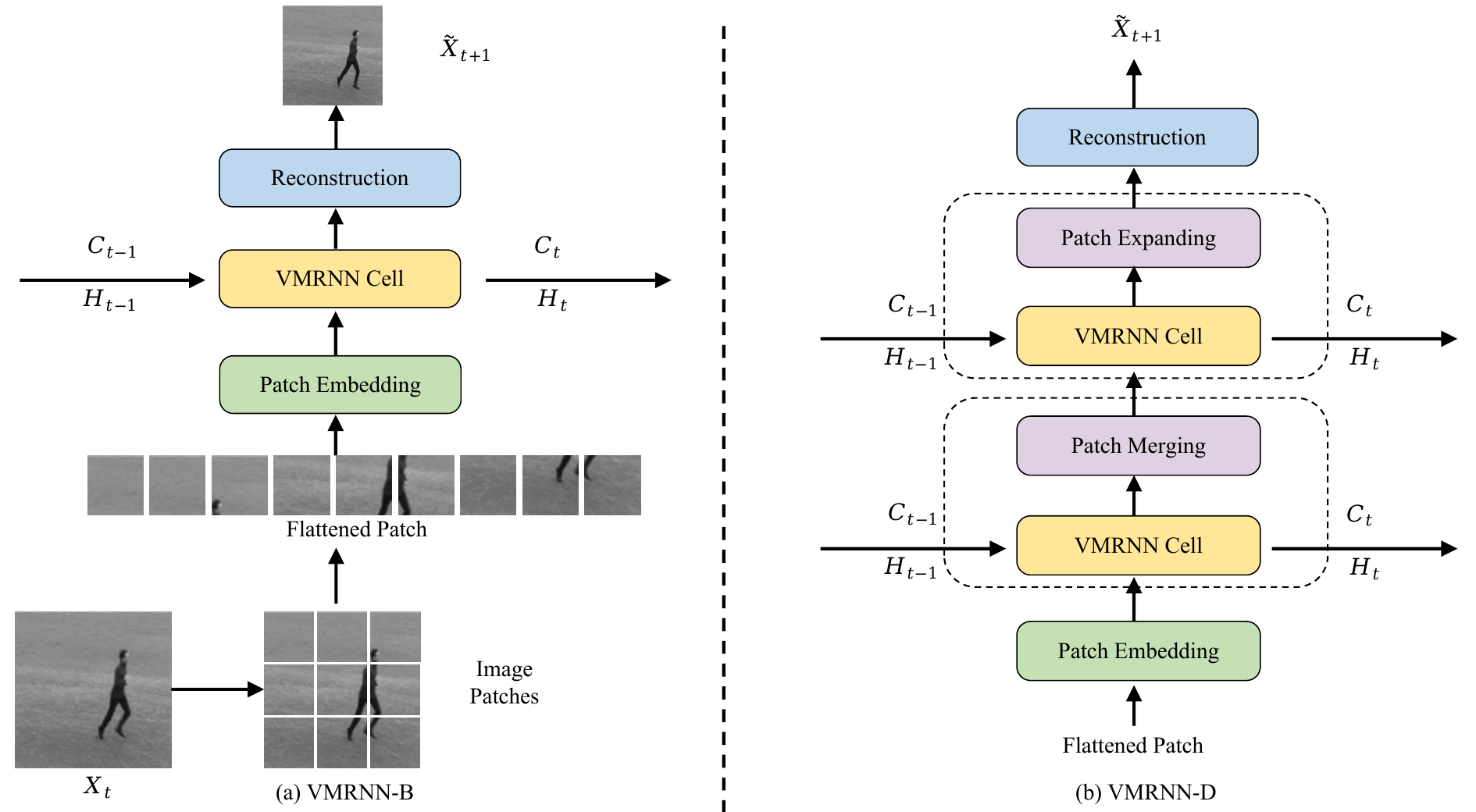}
	}
	\caption{(a): The architecture of the base model with a single VMRNN cell, VMRNN-B. (b): The architecture of the deep model with multiple VMRNN cells, VMRNN-D. }
	\label{fig:architecture}
\end{figure*}

\section{Method}
\subsection{Overall Architecture}
\label{sec:Architecture}

The architecture of our predictive model is illustrated in Fig.~\ref{fig:architecture} (a) and (b). Following the framework presented in ~\cite{Tang_2023_ICCV}, we introduce a base model and a deeper model centered on the VMRNN Cell, denoted as VMRNN-B and VMRNN-D, respectively. As shown in Fig.~\ref{fig:architecture} (a), at each time step, the image is divided into non-overlapping patches of size $P^{2}$ with patch size $P$. And then these image patches are flattened and sent into the patch embedding layer, which performs a linear transformation of the patches' original features into a specified dimensional space.

For the VMRNN-B model, the VMRNN layer processes the embedded image patches, along with the previous time step's hidden state $H_{t-1}$ and cell state ${C}_{t-1}$, to generate the current hidden state $H_{t}$ and cell state ${C}_{t}$. As illustrated in Fig.~\ref{fig:VMRNNCell}(a), $H_{t}$ is replicated, producing two versions: one is directed to the reconstruction layer, and the other, in conjunction with ${C}_{t}$, serves the VMRNN layer for the subsequent time step. For VMRNN-B, the architecture primarily relies on the stacking of VMRNN layers. For the VMRNN-D variant, we incorporate more VMRNN Cells and introduce Patch Merging and Patch Expanding layers, as outlined in ~\cite{cao2023swin}. The Patch Merging layer is employed for downsampling, effectively reducing the spatial dimensions of the data, which aids in reducing computational complexity and capturing more abstract, global features. Conversely, the Patch Expanding layer is used for upsampling, which increases the spatial dimensions, facilitating the restoration of detail and enabling precise localization of features in the reconstruction phase. Ultimately, the reconstruction layer takes the hidden state $H_{t}$ from the VMRNN layer and scales it back to the input size, generating the predicted frame for the next time step. 

The integration of downsampling and upsampling processes presents significant advantages in our predictive architecture. Downsampling simplifies the input representation, allowing the model to process higher-level features with reduced computational overhead. This is particularly beneficial for understanding complex patterns and relationships within the data at a more abstract level. Upsampling, on the other hand, ensures that the detailed spatial information is not lost. This balance between abstraction and detail preservation is key to achieving high-quality predictions, especially in tasks requiring fine-grained understanding and visual data generation.

\subsection{VMRNN Module}
\label{sec:VMRNN}


VMRNN  Module removes all weights $W$ and biases $b$ in ConvLSTM\cite{shi2015convolutional} to obtain Eqn.~\ref{eq:eq2}:

\begin{align}
  i_{t} &= f_{t} = o_{t} = \sigma\left(X_{t}+ H_{t-1}\right) \\
   {C}_{t} &=f_{t} \odot {C}_{t-1}+i_{t} \odot \tanh \left({X}_{t}+ {H}_{t-1}\right) \\
 H_{t} &=o_{t} \odot \tanh \left({C}_{t}\right)
	\label{eq:eq2}
\end{align}


We propose the VMRNN  Module, detailed in Fig.~\ref{fig:VMRNNCell} (a).  In VMRNN, the long-term and short-term temporal dependencies are captured by updating the information of cell states ${C}_{t}$, and hidden states $H_{t}$ are updated from a horizontal perspective. And the VSS Blocks vertically learn spatial dependencies. We show the key equations of VMRNN in Eqn.~\ref{eq:eq3}, where VSB means the VSS Blocks in Sec. \ref{sec:VSB} and LP is short for the Linear Projection:

\begin{align}
  {F}_{t} & =\sigma\left(\operatorname{VSB}\left(\operatorname{LP}\left(X_{t} ;  H_{t-1}\right)\right)\right) \\
  {C}_{t} & ={F}_{t} \odot\left(\tanh \left(\operatorname{VSB}\left(\operatorname{LP}\left(X_{t} ;  H_{t-1}\right)\right)\right)+{C}_{t-1}\right) \\
  H_{t} & ={F}_{t} \odot \tanh \left({C}_{t}\right)
\label{eq:eq3}
\end{align}

\subsection{VSS Block}
\label{sec:VSB}

The structure of VSS block is illustrated in Fig.~\ref{fig:VMRNNCell} (b). The process begins with the input being processed through an initial linear embedding layer, which is then split into two distinct information streams. The first stream is channeled through a $3\times 3$ depth-wise convolution layer, enriched with a Silu activation function~\cite{shazeer2020glu} before it progresses into the SS2D module. After SS2D, its output is refined by a layer normalization process, and subsequently, it merges with the second stream's output, which has been previously activated by Silu. This combination produces the final output of the VSS block. The architecture takes a novel path compared to vision transformer design, which typically follows a \texttt{Norm} $\rightarrow$ \texttt{attention} $\rightarrow$ \texttt{Norm} $\rightarrow$ \texttt{MLP} sequence within a block, and omits the \texttt{MLP} stage. This deviation renders the VSS block less complex than the ViT block, enabling the incorporation of a greater number of blocks within a comparable total model depth constraint.

VSS block first recovers linear projections to the image shape. (From [B, L, C] to [B, H, W, C]). Then VSS block addresses the challenges associated with 2D image data by employing 2D-selective-scan (SS2D), as illustrated in Fig.~\ref{fig:VMRNNCell} (c). This approach unfolds image patches in four distinct directions: from the top-left to the bottom-right, from the bottom-right to the top-left, from the top-right to the bottom-left, and from the bottom-left to the top-right, creating four distinct sequences, as depicted in the Scan Expand Stage. Subsequently, each feature sequence(scan) will be processed through the S6 Block. Finally, these sequences are reconfigured back into individual images, as depicted in the Scan Merge Stage. Given input feature $z$, the output feature $\Bar{z}$ of SS2D can be written as:
\begin{align}
    z_v &= expand(z, v) \\
    \Bar{z}_v &= S6(z_v) \label{eq:s6}\\
    \Bar{z} &= merge(\Bar{z}_1,\Bar{z}_2,\Bar{z}_3,\Bar{z}_4) 
\end{align}

where $v \in V=\{1,2,3,4\}$ is four different scanning directions. $expand(\cdot)$ and $merge(\cdot)$ corresponding to the \textit{scan expand} and \textit{scan merge} operations. The selective scan space state sequential model (S6) in Eqn. \ref{eq:s6} is the core SSM operator of the VSS block. It enables each element in a 1D array to interact with any of the previously scanned samples through a compressed hidden state. We plot the equations of S6 process in Fig.~\ref{fig:VMRNNCell} (c). in S6 Block stage.

\section{Experiments}
\label{sec:Experiments}

\subsection{Implementations}
We employ the Mean Squared Error (MSE) loss function across all three datasets. For KTH~\cite{schuldt2004recognizing} and TaxiBJ~\cite{zhang2017deep}, our methodology aligns with OpenSTL~\cite{tan2024openstl}. For the Moving MNIST~\cite{srivastava2015unsupervised} dataset, we adhere to the experimental setup detailed in ~\cite{Tang_2023_ICCV}. The precise model parameters, hyper-parameters(including batch size, learning rate, and training epochs), and training machines utilized for each dataset are comprehensively enumerated in Table~\ref{tab:para}. For TaxiBJ, we train 200 epochs with a learning rate of 4e-4 with a single A6000 GPU, using a batch size of 16. For KTH, we train 100 epochs with a learning rate of 5e-4 and 1e-4 for KTH20 and KTH40, respectively. For the Moving MNIST dataset, we adhere to the experimental setup detailed in ~\cite{Tang_2023_ICCV} and train 2000 epochs with a learning rate of 5e-5 and a batch size of 8 using a single RTX 3090Ti GPU.

We utilize an extensive array of evaluation metrics, including Mean Squared Error (MSE), Mean Absolute Error (MAE), Peak Signal Noise Ratio (PSNR), and the Structural Similarity Index Measure (SSIM)~\cite{wang2004image}. These metrics are computed across all predicted frames, where lower MAE and MSE scores, or higher SSIM and PSNR scores, signify superior prediction precision. To assess the models' computational demand, we measure the number of parameters, and floating-point operations (FLOPs) on TaxiBJ, and report the inference speed in frames per second (FPS) on a single NVIDIA A6000 GPU. This multifaceted evaluation provides insight into the efficiency and scalability of different models.




Following SwinLSTM~\cite{Tang_2023_ICCV}, we adopt MSE and SSIM as our metrics for evaluating the Moving MNIST dataset, and SSIM and PSNR for the KTH dataset. Following OpenSTL~\cite{tan2024openstl}, we provide a comprehensive analysis of the TaxiBJ dataset that includes not just MSE, MAE, and SSIM, but also detailed evaluations of model parameters and computational complexity, measured in FLOPs.


\begin{figure*}[h]
	\centering
	\resizebox{\textwidth}{!}{
	\includegraphics{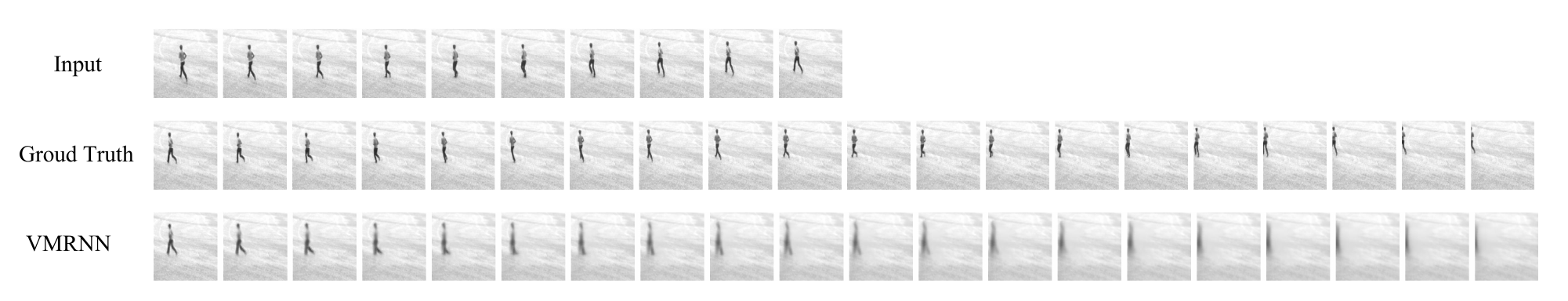}
	}
	\caption{Qualitative results of VMRNN on KTH.}
	\label{fig:kth_vis}
\end{figure*}

We chose three pivot datasets across various domains, including synthetic moving object trajectory, human motion, and traffic flow.

\noindent{\bf Moving MNIST.}  The moving MNIST dataset~\cite{srivastava2015unsupervised} serves as a benchmark synthetic dataset for evaluating sequence prediction models. Our approach to generating Moving MNIST sequences is in line with the methodology described in ~\cite{srivastava2015unsupervised}, where each sequence comprises 20 frames. We designate the initial 10 frames for input and the subsequent 10 frames as the prediction target. We adopt 10000 sequences for training and for fair comparisons, we use the pre-generated 10000 sequences~\cite{gao2022simvp} for validation.


\noindent{\bf KTH.}  The KTH dataset~\cite{schuldt2004recognizing} contains 25 individuals performing six types of human actions (walking, jogging, running, boxing, hand waving, and hand clapping)  in 4 different scenarios. Following previous works~\cite{wang2018eidetic,gao2022simvp,tan2024openstl}, we use persons 1-16 for training and persons 17-25 for validation and resizing each image to $128\times128$. The models predict 10 frames from 10 observations at training time and 20 or 40 frames at inference time.

\noindent{\bf TaxiBJ.}  TaxiBJ~\cite{zhang2017deep} includes GPS data from taxis and meteorological data in Beijing. Each data frame is visualized as a $32\times32\times2$ heatmap, where the third dimension encapsulates the inflow and outflow of traffic within a designated area. Adhering to the experimental framework proposed in ~\cite{zhang2017deep}, we allocate the final four weeks of data for testing, while the preceding data is utilized for training. Our prediction model is tasked with using four sequential observations to forecast the subsequent four frames.

\begin{table*}[h]
	\centering
 	\caption{Experimental setup. $\textbf{VMRNNs}$ denotes the number of the VMRNN Cells in the spatio-temporal forecasting network. $\textbf{VSB}$ denotes the number of the VSS Blocks in VMRNN cell. $\textbf{Patch size}$ indicates the patch token size. }
	\label{tab:para}
        \scriptsize
	$
	\begin{array}{cccccccccccc}
		\toprule \text { Dataset }& \text { Model }& \text { VMRNNs } & \text { VSB } & \text { Patch Size } & \text { Resolution } & \text { Train } & \text { Test }& \text { Epochs }& \text { Learning Rate }& \text { Batch Size }& \text { GPU }\\
		\midrule 
		\text {Moving MNIST} & \text {VMRNN-D}& \text {4}& \text {(2, 6, 6, 2)} & \text{2} & \text{(64, 64, 1)} & \text {10} \rightarrow \text {10} & \text {10} \rightarrow \text {10}& \text {2000}& \text {5e-5}& \text {8}& \text {1 * RTX 3090}\\
		\text {KTH}& \text {VMRNN-B}& \text {1} & \text{6} & \text{2}  & \text{(128, 128, 1)} & \text {10}\rightarrow \text {10} & \text {10}\rightarrow \text {20/40}& \text {100}& \text {5e-4/1e-4}& \text {2/1}& \text {4 * A6000}\\
		\text {TaxiBJ}& \text {VMRNN-B}& \text {1} & \text {12} & \text{4}  & \text{(32, 32, 2)} & \text {4} \rightarrow \text {4}& \text {4}\rightarrow \text {4} & \text {200}& \text {4e-4}& \text {16}& \text {1 * A6000}\\
		\bottomrule
	\end{array}
	$

\end{table*}

\subsection{Main results}

Tables~\ref{tab:mm}, \ref{tab:taxibj}, and \ref{tab:KTH} provide quantitative comparisons between VMRNN and previous state-of-the-art models across three distinct datasets. These comparisons highlight VMRNN's exceptional capability as an efficient and highly generalizable approach for spatiotemporal prediction.

\noindent{\textbf {Moving MNIST}}
We present the quantitative outcomes in Table~\ref{tab:mm}, where our VMRNN model demonstrates notably superior performance compared to all other evaluated models. We report the results of previous research directly. On the Moving MNIST dataset, VMRNN not only achieves obviously lower MSE but also secures higher SSIM scores, significantly surpassing earlier methods by a substantial margin and archives 6.8\% improvement over SwinLSTM.

\begin{table}
	\centering
 	\caption{ Quantitative comparison of VMRNN and other methods on \textbf{Moving MNIST}. Each model observes 10 frames and predicts the subsequent 10 frames. Lower MSE and higher SSIM indicate better predictions.}
        $
	\begin{array}{ccc}
		\toprule \text { Method }  & \text { MSE } \downarrow & \text { SSIM }\uparrow  \\
		\midrule 
		\text { ConvLSTM~\cite{shi2015convolutional} }  & \text{103.3} & \text{0.707} \\ 
		\text { DFN~\cite{jia2016dynamic} }  & \text{89.0} & \text{0.726} \\ 
		\text { FRNN~\cite{oliu2018folded} }  & \text{69.7} & \text{0.813} \\ 
		\text { VPN~\cite{kalchbrenner2017video} }  & \text{64.1} & \text{0.870} \\
		\text { PredRNN~\cite{wang2017predrnn} }  & \text{56.8} & \text{0.867} \\
		\text { CausalLSTM~\cite{wang2018predrnn++} }  & \text{46.5} & \text{0.898} \\
		\text { MIM~\cite{wang2019memory} } & \text{44.2} & \text{0.910}  \\
		\text { E3D-LSTM~\cite{wang2018eidetic} }  & \text{41.3} & \text{0.910} \\
		\text { LMC~\cite{lee2021video} }  & \text{41.5} & \text{0.924} \\
		\text { MAU~\cite{chang2021mau} }  & \text{27.6} & \text{0.937} \\
		\text { PhyDNet~\cite{guen2020disentangling}}  & \text{24.4} & \text{0.947}\\
		\text { CrevNet~\cite{yu2020efficient}}  & \text{22.3} & \text{0.949}\\
            \text {SimVP~\cite{gao2022simvp}} & \text{23.8} & \text{0.958} \\
            \text {TAU~\cite{tan2023temporal}} & \text{19.8} & \text{0.957} \\
            \text {MMVP~\cite{Zhong_2023_ICCV}} & \text{22.2} & \text{0.948} \\
		\text {SwinLSTM~\cite{Tang_2023_ICCV}} & \text{17.7} & \text{0.962} \\ 
            \text {VMRNN } &\textbf{16.5} & \textbf{0.965} \\
		\bottomrule
	\end{array}
        $
	\label{tab:mm}
\end{table}

\noindent{\textbf {TaxiBJ}}
We present the quantitative outcomes in Table~\ref{tab:taxibj}, where our VMRNN model demonstrates notably superior performance compared to all other evaluated models. For SwinLSTM, which is not reported in OpenSTL, we follow the same hyper-parameters with our VMRNN to ensure a fair comparison. For other methods, we use the results reported in OpenSTL. Obviously, on the TaxiBJ dataset, VMRNN not only registers substantially lower MSE and MAE values but also attains higher SSIM scores, thereby eclipsing previous methodologies to a considerable extent.

In Fig.~\ref{fig:flops}, we provide a comparative analysis of parameters and FLOPs among recent spatiotemporal predictive learning methodologies applied to the TaxiBJ dataset. A position towards the lower left indicates a model that not only requires fewer parameters and computational resources but also delivers superior predictive performance, as evidenced by lower MSE values. Our VMRNN model showcases remarkable efficiency and effectiveness, achieving a clear lead by requiring fewer parameters and FLOPs while maintaining high prediction accuracy. As for Moving MNIST and KTH, the performance is similar, with fewer parameters and FLOPs than other methods.

\begin{table}[h]
  \vspace{-2mm}
  \small
  \centering
  \setlength{\tabcolsep}{1.5mm}
  \renewcommand\arraystretch{1.00}
  \caption{The performance on the TaxiBJ dataset. We provide a comparative analysis of parameters and FLOPs among recent spatiotemporal predictive learning methodologies. Our VMRNN model showcases remarkable efficiency and effectiveness.}
  \resizebox{0.5\textwidth}{!}{%
    \begin{tabular}{cccccccc}
      \toprule
       \multicolumn{2}{c}{Method} & Para.(M) & FLOPs (G) & FPS & MSE $\downarrow$ & MAE $\downarrow$ & SSIM $\uparrow$\\ \hline
      & ConvLSTM~\cite{shi2015convolutional} & 15.0 & 20.7 & 815 & 0.3358 & 15.32 & 0.9836 \\
      & PredNet~\cite{prednet} & 12.5 & 0.9 & 5031 & 0.3516 & 15.91 & 0.9828 \\
      & PredRNN~\cite{wang2017predrnn} & 23.7 & 42.4 & 416 & 0.3194 & 15.31 & 0.9838 \\
      & PredRNN++~\cite{wang2018predrnn++} & 38.4 & 63.0 & 301 & 0.3348 & 15.37 & 0.9834  \\
      & E3DLSTM~\cite{wang2018eidetic} & 51.0 & 98.19 & 60 & 0.3421 & 14.98 & 0.9842 \\
       & PhyDNet~\cite{guen2020disentangling} & 3.1 & 5.6 & 982 & 0.3622 & 15.53 & 0.9828  \\
       & MIM~\cite{wang2019memory} & 37.9 & 64.1 & 275 & 0.3110 & 14.96 & 0.9847 \\
       & MAU~\cite{chang2021mau} & 4.4 & 6.0 & 540 & 0.3268 & 15.26 & 0.9834 \\
       & PredRNNv2~\cite{predrnnv2}& 23.7 & 42.6 & 378 & 0.3834 & 15.55 & 0.9826  \\
       & SimVP~\cite{gao2022simvp} & 13.8 & 3.6 & 533 & 0.3282 & 15.45 & 0.9835 \\
       & TAU~\cite{tan2023temporal} & 9.6 & 2.5 & 1268 & 0.3108 & 14.93 & 0.9848 \\
        & SimVPv2~\cite{tan2022simvpv2} & 10.0 & 2.6 & 1217 & 0.3246 & 15.03 & 0.9844 \\
        & SwinLSTM~\cite{Tang_2023_ICCV}  &  2.9  &  1.3  & 1425 &  0.3026 &  15.00& 0.9843    \\
      & VMRNN & 2.6 & 0.9 & 526 & \textbf{0.2887} & \textbf{14.69} & \textbf{0.9858}  \\ \bottomrule
    \end{tabular}%
  }
  \label{tab:taxibj}
  \vspace{-2mm}
\end{table}

\noindent{\textbf {KTH}}
We present the quantitative results in Table~\ref{tab:KTH}. Our VMRNN model shows higher SSIM than all previous methods and comparable PSNR value with SwinLSTM. We report the results from the previous study directly and from our practice in OpenSTL, VMRRN achieves both better results in PSNR and SSIM than SwinLSTM by a large margin, either following the hyper-parameter setting in SwinLSTM or adopting the same setting as VMRNN.

\begin{table}
	\centering
 	\caption{Quantitative evaluation on the \textbf{KTH} test set. We present the model observing 10 frames to predict 20 or 40 frames, and all metrics are averaged over the predicted frames. Higher SSIM and PSNR indicate better prediction quality.}
	\resizebox{\hsize}{!}{
		$
		\begin{array}{ccccc}
			\toprule & \multicolumn{2}{c}{\text { KTH }(\text {10}\rightarrow\text {20})} & \multicolumn{2}{c}{\text { KTH }(\text {10}\rightarrow\text {40})} \\
			\cline { 2 - 5 } \text { Method } & \text { SSIM }\uparrow & \text { PSNR }\uparrow & \text { SSIM }\uparrow & \text { PSNR }\uparrow \\
			\midrule
			\text { ConvLSTM~\cite{shi2015convolutional} } &\text {0.712} & \text {23.58} & \text {0.639} & \text {22.85} \\
			\text { SAVP~\cite{gao2019disentangling} } &\text {0.746} & \text {25.38} & \text {0.701} & \text {23.97} \\
			\text { FRNN~\cite{oliu2018folded} } & \text {0.771} & \text {26.12} &\text {0.678} & \text {23.77}\\
			\text { DFN~\cite{jia2016dynamic} } & \text {0.794} & \text {27.26}  & \text {0.652} & \text {23.01} \\
			\text { PredRNN~\cite{wang2017predrnn}  } & \text {0.839} & \text {27.55} & \text {0.703} & \text {24.16} \\
			\text { VarNet~\cite{jin2018varnet}  } & \text {0.843} & \text {28.48} & \text {0.739} & \text {25.37} \\
			\text { SVAP-VAE~\cite{lee2018stochastic}  } & \text {0.852} & \text {27.77} & \text {0.811} & \text {26.18} \\
			\text { PredRNN++~\cite{wang2018predrnn++}  } & \text {0.865} & \text {28.47} & \text {0.741} & \text {25.21} \\
			\text { E3d-LSTM~\cite{wang2018eidetic}  } & \text {0.879} & \text {29.31} & \text {0.810} & \text {27.24} \\
			\text { STMFANet~\cite{jin2020exploring}  } & \text {0.893} & \text {29.85} & \text {0.851} & \text {27.56} \\
                \text {SwinLSTM~\cite{Tang_2023_ICCV} } & \text{0.903} & \textbf{34.34} & \text{0.879} & \textbf{33.15} \\
                 \text {VMRNN} & \textbf{0.907} & \text{34.06} & \textbf{0.882} & \text{32.69} \\
			\bottomrule
		\end{array}
		$
	}	
	\label{tab:KTH}
\end{table}

\subsection{Ablation Study}

In this section, we perform ablation studies on TaxiBJ to analyze the impact of different elements on model performance. We discuss three major elements: the convolution layer, patch size, and the number of VSS Blocks.

\noindent{\bf The Convolution Layer.} 
The role of the VSS Block convolution layer is to decode the spatiotemporal representations extracted by the VMRNN cell. We conduct experiments on depth-wise convolutions (DW Conv), convolution, and depth-wise convolutions with dilations (DW-D Conv) to analyze the impact of different decoding methods. Following ~\cite{tan2023temporal}, we combine DW Conv-DW-D Conv-1×1 Conv to model the large kernel convolutions, Table~\ref{tab:convolution} shows that DW Conv performs much better than the other two methods.

\noindent{\bf Patch Size.} 
The choice of image patch size critically influences the length of the input token sequences, where smaller patch sizes yield longer sequences. To evaluate the impact of different patch sizes on performance, we conducted experiments on the TaxiBJ and the KTH dataset using patch sizes 2, 4, and 8. As depicted in Table~\ref{tab:Patch}, a patch size of 4 for TaxiBJ and 2 for KTH distinctly outperforms the others.

\begin{table}[h]
    \centering
        \centering
        \caption{Ablation study on patch size of TaxiBJ and KTH.}
        \small
        $
        \begin{array}{c|ccc|cc}
            \toprule
            &\multicolumn{3}{c|}{\text { TaxiBJ}}  &\multicolumn{2}{c}{\text {KTH 10} \rightarrow \text {20}}\\
            \text {Patch size}& \text { MSE } & \text { MAE }& \text { SSIM }& \text { SSIM }& \text { PSNR }  \\
            \midrule 
            \text { 2 } & \text {0.3566} & \text {15.46} & \text {0.9823} & \textbf {0.907} & \textbf {34.06} \\
            \text { 4 } & \textbf{0.2887} & \textbf{14.69}  & \textbf{0.9858} & \text {0.887} & \text {32.87}\\
            \text { 8 } & \text {0.3400} & \text {15.32}  & \text {0.9843} & \text {0.874} & \text {32.10}\\
            \bottomrule
        \end{array}
        $
        \label{tab:Patch}
\end{table}

\begin{table}[h]
    \centering
        \caption{Ablation study on different convolution methods of VSS Block on TaxiBJ. }
        \small
	$
	\begin{array}{c|ccc}
		\toprule
		&\multicolumn{3}{c}{\text { TaxiBJ}}  \\
		\text {Convolution Method} & \text { MSE } & \text { MAE }& \text { SSIM }  \\
		\midrule 
		\text { DWConv }  & \textbf {0.2887} & \textbf {14.69} & \textbf {0.9858}  \\
		\text { Conv2d }   & \text{0.3185} & \text{16.58}  & \text {0.9797 } \\
		\text { DW-DW-D-1x1} & \text {0.3050} & \text {15.02}  & \text {0.9853} \\
		\bottomrule
	\end{array}
	$
	\label{tab:convolution}
\end{table}

\begin{figure}[h]
    \centering
    \begin{tabular}{cc}
    \hspace{-0.5cm} \includegraphics[width=0.45\columnwidth]{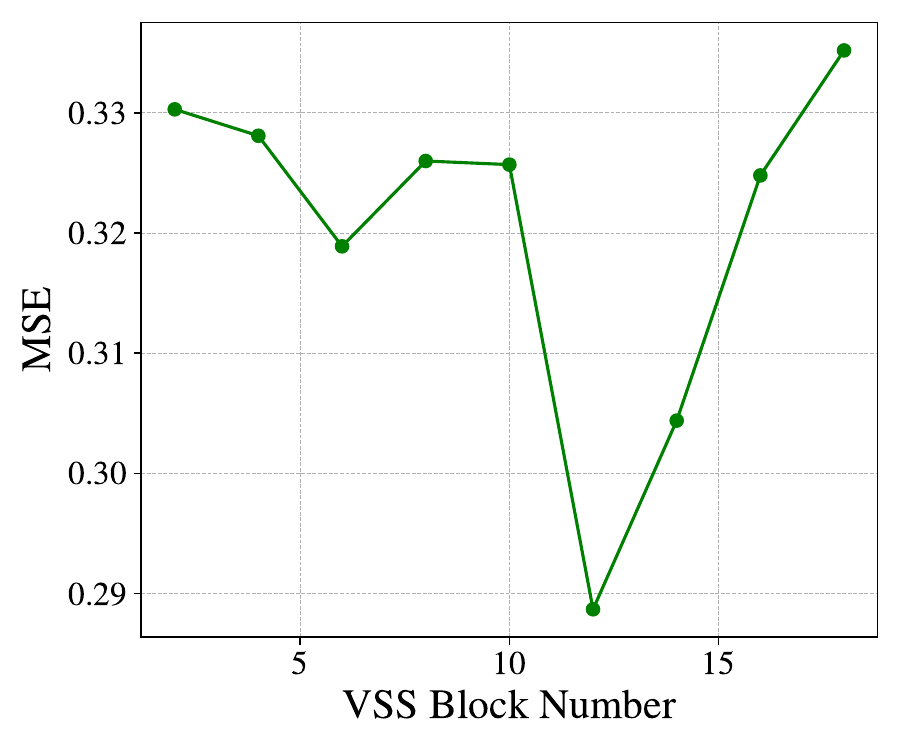} &
    \hspace{-0.3cm} \includegraphics[width=0.45\columnwidth]{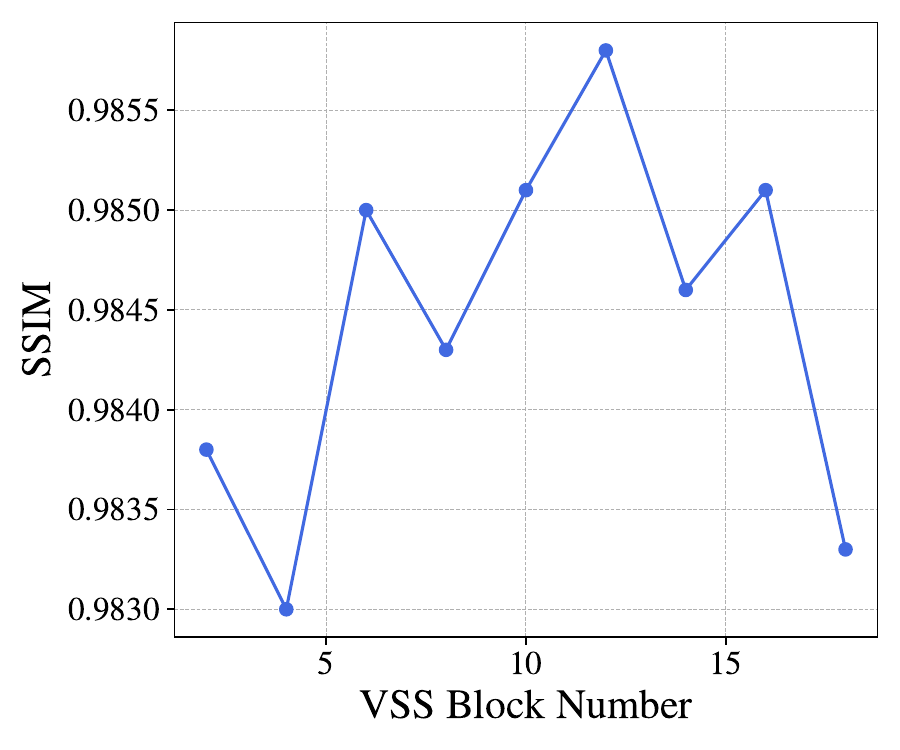}
    \\(a) MSE & (b) SSIM 
    \end{tabular} \\
    \caption{Ablation study on the different numbers of VSS Block with VMRNN on TaxiBJ.}
    \label{fig:line_chart}
\end{figure}

\noindent{\bf The Number of VSS Blocks.} 
To investigate the capabilities of different VSS blocks to model global spatial information, we examined the impact of varying the number of VSS blocks from 2 to 18. Fig.~\ref{fig:line_chart} illustrates the MSE and SSIM outcomes across different counts of VSS blocks on the TaxiBJ dataset. The results exhibit a trend of improvement followed by deterioration as the number of blocks increases, with an optimal performance observed at 12 VSS blocks, which achieve the best results in terms of both MSE and SSIM metrics.

\subsection{Visualization}

\begin{figure}[ht]
	\centering
	\resizebox{0.45\textwidth}{!}{
	\includegraphics{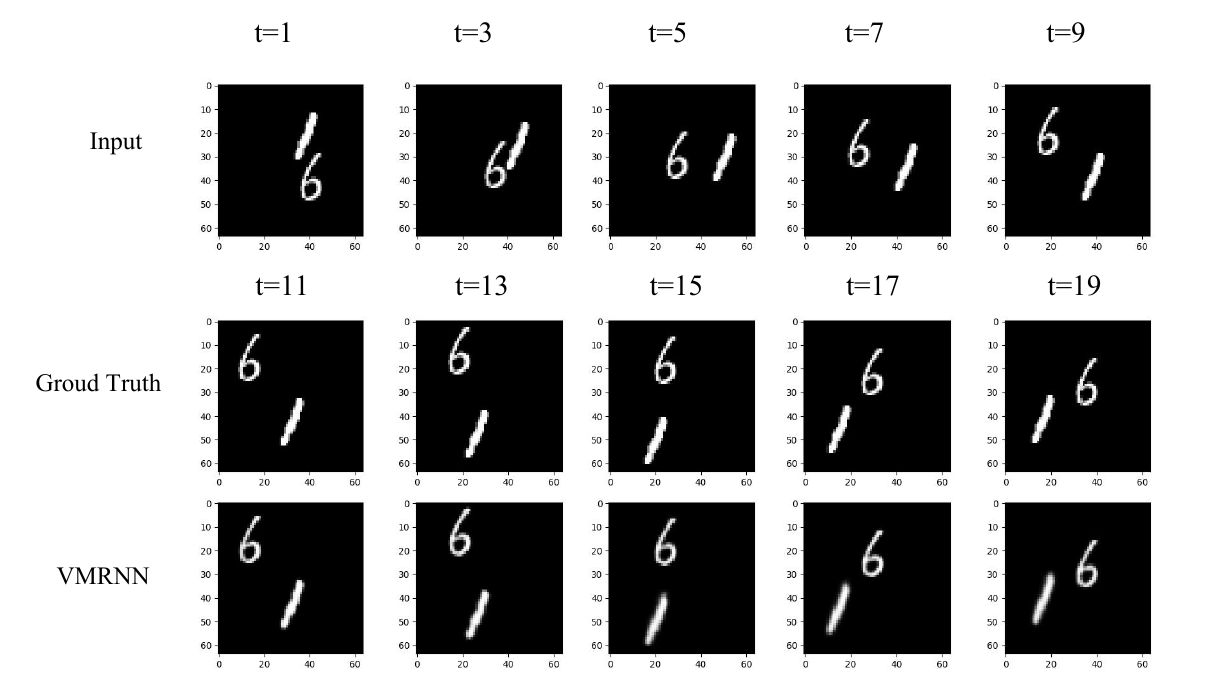}
	}
	\caption{Qualitative results of VMRNN on Moving MNIST.}
	\label{fig:mm_vis}
\end{figure}

\begin{figure}[h]
	\centering
	\resizebox{0.45\textwidth}{!}{
	\includegraphics{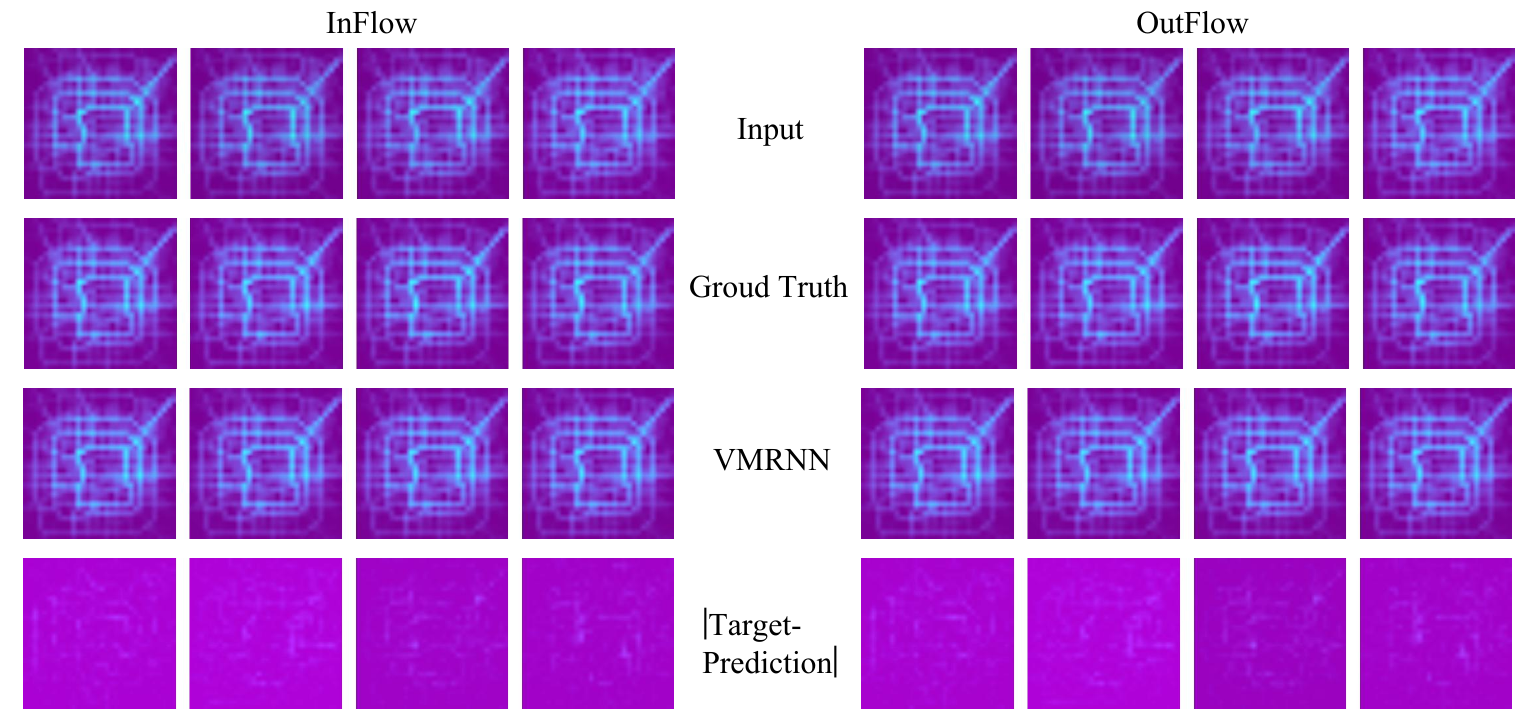}
	}
	\caption{Qualitative results of VMRNN on TaxiBJ.}
	\label{fig:taxibj_vis}
\end{figure}

We present qualitative results of VMRNN on Moving MNIST in Fig.~\ref{fig:mm_vis}, TaxiBJ in Fig.~\ref{fig:taxibj_vis}, and KTH in Fig.~\ref{fig:kth_vis}. For all datasets, the first line is the input, the second line is the ground truth, and the third line is the prediction of VMRNN. For TaxiBJ, we add the fourth line to show the difference between prediction and target. The visualization results reveal that VMRNN delivers impressive predictive performance, maintaining high accuracy across extended horizons.


\section{Conclusion}

In this work, we introduce VMRNN, a novel approach that integrates LSTM architecture with VSS Blocks to tackle spatiotemporal forecasting challenges. Through rigorous evaluation across diverse datasets, VMRNN has proven its prowess by delivering superior performance while maintaining a smaller model size. This advancement is attributed to the model's enhanced capability to learn and leverage global spatial dependencies with linear complexity, enabling a more refined understanding of spatiotemporal dynamics. The findings show that VMRNN sets a new strong baseline for future explorations in the field.

\section{Acknowledgements}

This work was supported by the National Natural Science Foundation of China (No. 62306257). The views and conclusions contained herein are those of the authors and should not be interpreted as necessarily representing the official policies or endorsements, either expressed or implied, of the National Natural Science Foundation.

\par\vfill\par

\clearpage  

{\small
\bibliographystyle{ieee_fullname}
\bibliography{main}
}

\end{document}